\documentclass[journal]{IEEEtran}
\usepackage{cite}
\usepackage{bm}
\usepackage{graphicx}
\usepackage{caption}
\usepackage{mathrsfs}
\usepackage{amsmath}
\usepackage{array}
\usepackage[table]{xcolor}
\usepackage{color}
\usepackage{epsfig,epstopdf,subfigure}

\renewcommand{\arraystretch}{1.3}

\begin{document}
	
	\title{Phonetic Temporal Neural Model for Language Identification}

	\author{Zhiyuan Tang, Dong~Wang,~\IEEEmembership{Senior Member,~IEEE}, Yixiang~Chen, Lantian~Li, and Andrew Abel % <-this % stops a space
		
		\thanks{
%			Manuscript received XXXX, 2017; revised XXXX, 2017; accepted
%			XXXX, 2017. Date of publication XXXX, 2017; date of current
%			version XXXX, 2017.
			This work was supported in part by the National Natural Science
			Foundation of China under Projects 61271389, 61371136, and 61633013,
			in part by the National Basic Research Program (973 Program) of China under
			Grant 2013CB329302, in part by the Doctoral Fund of Ministry of Education
			of China under Project 20130002120011.
%			The associate editor coordinating the review of this manuscript and
%			approving it for publication was XXXX.
			(\emph{Corresponding author: Dong Wang.})
			
			Z. Tang is with the Chengdu Institute of Computer Applications,
			Chinese Academy of Sciences and University of Chinese Academy of Sciences, Beijing 100049, China,
			and also with the Center for Speech and Language Technologies, Tsinghua University, Beijing 100084, China
			(e-mail: tangzy@cslt.riit.tsinghua.edu.cn).
			
			D. Wang, Y. Chen and L. Li are with the Tsinghua National Laboratory for
			Information Science and Technology and the Center for Speech and Language
			Technologies, Tsinghua University, Beijing 100084, China
			(e-mail: wangdong99@mails.tsinghua.edu.cn; \{chenyx,lilt\}@cslt.riit.tsinghua.edu.cn).
			
			A. Abel is with the Department of Computer Science and Software Engineering, Xi'an Jiaotong-Liverpool University, Suzhou 215123, China (andrew.abel@xjtlu.edu.cn).
			
		}
	}

	\maketitle
	
	\begin{abstract}

		Deep neural models, particularly the LSTM-RNN model, have shown great potential for language identification (LID). However, the use of phonetic information has been largely overlooked by most existing neural LID methods, although this information has been used very successfully in conventional phonetic LID systems.  We present a phonetic temporal neural model for LID, which is an LSTM-RNN LID system that accepts phonetic features produced by a phone-discriminative DNN as the input, rather than raw acoustic features.  This new model is similar to traditional phonetic LID methods, but the phonetic knowledge here is much richer: it is at the frame level and involves compacted information of all phones. Our experiments conducted on the Babel database and the AP16-OLR database demonstrate that the temporal phonetic neural approach is very effective, and significantly outperforms existing acoustic neural models. It also outperforms the conventional i-vector approach on short utterances and in noisy conditions.

	\end{abstract}
	
	\begin{IEEEkeywords}
		Language identification; Deep neural networks; Multi-task learning
	\end{IEEEkeywords}
	
	% \fi
	\IEEEpeerreviewmaketitle

	\section{Introduction}
	\label{sec:intro}
	
	Language identification (LID) lends itself to a wide range of applications, such as mixed-lingual (code-switching) speech recognition. Humans use many cues to discriminate languages, and better accuracy can be achieved with the use of more cues. Various LID approaches have been developed, based on different types of cues.
	
	\subsection{Cues for language identification}
	
	There are more than 5000 languages in the world, and each language has distinct properties at different levels, from acoustic to semantics~\cite{fromkin2010introduction,comrie2009world,crystal2cambridge}.  A number of studies have investigated how humans use these properties as cues to distinguish between languages~\cite{haper07}. For example, Muthusamy~\cite{muthusamy1994perceptual} found that familiarity with a language is an important factor affecting LID accuracy, and that longer speech samples are easier to identify. Moreover, people can easily tell what cues they use for identification, including phonemic inventory, word usage, and prosody. More thorough investigations were conducted by others by modifying speech samples to promote one or several factors. For example, Mori et al. ~\cite{mori1999human} found that people are able to identify Japanese and English fairly reliably even when phone information is reduced. They argued that other non-linguistic cues such as intensity and pitch were used to decide the language. Navratil~\cite{navratil2001spoken} evaluated the importance of various types of knowledge, including lexical, phonotactic and prosodic, by asking humans to identify five languages, Chinese, English, French, German and Japanese.  Subjects were presented with unaltered speech samples, samples with randomly altered syllables, and samples with the vocal-tract information removed to leave only the F0 and amplitude. Navratil found that the speech samples with random syllables are more difficult to identify compared to the original samples (73.9\% vs 96\%), and removing vocal-tract information leads to significant performance reduction (73.9\% vs 49.4\%). This means that with this 5-language LID task, the lexical and phonotactic information is important for human decision making.
	
	The LID experiments summarised above suggest that languages can be discriminated by multiple cues at different levels, and the cues used to differentiate different language pairs are different. In general, the cues can be categorized into three levels: feature level, token level and prosody level.
	At the feature level, different languages have their own implementation of phones, and the transitions between phones are also different. This acoustic speciality is a short-time property and can be identified by certain spectral analysis and feature extraction of our auditory system.  At the token level, the distribution and transition patterns of linguistic tokens at various levels are significantly different. The tokens can be phones/phonemes, syllables, words or even syntactic or semantic tags.  At the prosody level, the duration, pitch and stress patterns often differ between languages. For example, patterns of stress can provide an important cue for discriminating between two stressed languages, duration can also be potentially useful, and the tone patterns of syllables or words offer a clear cue to discriminate between tonal languages.

	\subsection{LID approaches}
	
	Based on the different types of cues, multiple LID approaches have been proposed.  Early work generally focused on feature-level cues.  Feature-based methods use strong statistical models built on raw acoustic features to make the LID decision.  For instance, Cimarusti used LPC features~\cite{cimarusti1982development}, and Foil et al.~\cite{foil1986language} investigated formant features. Dynamic features that involve temporal information were also demonstrated to be effective~\cite{torres2002approaches}.  The statistical models used include Gaussian mixture models (GMMs)~\cite{zissman1993automatic,willmore2000comparing}, hidden Markov models (HMMs)~\cite{wong2004automatic,nakagawa1992speaker}, neural networks (NNs)~\cite{kwasny1992identifying,muthusamy1993segmental}, and support vector machines (SVMs)~\cite{campbell2004language}. More recently, a low-rank GMM model known as the i-vector model was proposed and achieved significant success~\cite{Najim2011lang,martinez2011language}. This model constrains the mean vectors of the GMM components in a low-dimensional space to improve the statistical strength for model training, and uses a task-oriented discriminative model (e.g., linear discriminative analysis, LDA) to improve the decision quality at run-time, leading to improved LID performance.  Due to the short-time property of the features, most feature-based methods model the distributional characters rather than the temporal characters of speech signals.

	The token-based approach is based on the characters of high-level tokens. Since the dynamic properties of adjacent tokens are more stable than adjacent raw features, temporal characters can be learned with the token-based approach, in additional to the distributional characters.  A typical approach is to convert speech signals into phone sequences, and then build an n-gram language model (LM) for each target language to evaluate the confidence that the input speech matches that language. This is the famous phone recognition and language modelling (PRLM) approach. Multiple PRLM variants have been proposed, such as parallel phone recognition followed by LM (PPRLM)~\cite{zissman1996comparison,matejka2006brno}, and phone recognition on a multilingual phone set~\cite{hazen1997segment}.  Other tokens such as syllables~\cite{zhu2005different} and words~\cite{schultz1996lvcsr,hieronymus1997robust} have also been investigated.
	
	The prosody-based approach utilizes patterns of duration, pitch, and stress to discriminate between languages. For example, Foil et al.~\cite{foil1986language} studied formant and prosodic features and found formant features to be more discriminative. Rouas et al.~\cite{rouas2003modeling} modeled pure prosodic features by GMMs and found that their system worked well on read
	speech, but could not deal with the complexity of spontaneous speech prosody.
	%[1]
	Muthusamy~\cite{muthusamy1993segmental} used pitch variation, duration and syllable rate.  Duration and pitch patterns were also used by Hazen~\cite{hazen1997segment}.  In most cases, the prosodic information is used as additional knowledge to improve feature or token-based LID.
	
	Most of the above methods, no matter what information is used, heavily rely on probabilistic models to accumulate evidence from a long speech segment. For example, the PRLM method requires an n-gram probability of the phonetic sequence, and the GMM/i-vector method requires the distribution of the acoustic feature. Therefore, these approaches require long test utterances, leading to inevitable latency in the LID decision.
	This latency is a serious problem for many practical applications, e.g.,
	code-switching ASR, where multiple languages may be contained within a single block of speech. For quick LID, frame-level decision is highly desirable, which therefore cannot rely on probabilistic models.
	
	The recently emerging deep learning approach solves this problem by using various deep neural networks (DNNs) to produce frame-level LID decisions.  An early successful deep neural model was developed by Lopez-Moreno et al.~\cite{lopez2014automatic}, who proposed an approach based on a feed-forward deep neural network (FFDNN), which accepts raw acoustic features and produces frame-level LID decisions. The score for utterance-based decision is calculated by averaging the scores of the frame-level decisions.  This was extended by others with the use of various neural model structures, e.g., CNN~\cite{lozano2015end,jin2016lid} and TDNN~\cite{kotov2016language,garcia2016stacked}.  These DNN models are feature-based, but they consider a large context window, and can therefore learn the feature's temporal information, which is not possible with conventional feature-based models (such as the i-vector model), that only learn distributional information.  The temporal information can be better learned by recurrent neural networks (RNNs), as proposed by Gonzalez-Dominguez et al.~\cite{gonzalez2014automatic}. Using an RNN structure based on the long-short term memory unit (LSTM), the authors reported better performance with fewer parameters. This RNN approach was further developed by others, e.g.,~\cite{gelly2016divide,zazo2016language}.
	
	%Due to the frame-level decision, the neural approach tends to be more superior with short utterances(e.g., 2$\sim$3 seconds)
	%compared to the probabilistic methods. This has been demonstrated by a number of researchers, e.g.,~\cite{lopez2014automatic,gonzalez2014automatic,zazo2016language}.
	
	%We note that
	%the frame-based decision using neural models was considered by researchers much earlier before DNN getting
	%popularity~\cite{willmore2000comparing}, but it did not achieve much success until the development of
	%deep learning. This is probably because learning language-related information from raw acoustic features is highly challenging.
	%Deep learning offers such a possibility by discovering language-related information by hierarchical feature learning.
	
	It should be noted that DNNs have been used in other ways in LID. For example, Song et al.~\cite{song2013vector} used a DNN to extract phonetic feature for the i-vector system, and Ferrer et al.~\cite{ferrer2016study} proposed a DNN i-vector approach that uses posteriors produced by a phone-discriminative FFDNN to compute the Baum-Welch statistics. Tian et al.~\cite{tian2016investigation} extended this by using an RNN to produce the posteriors. These methods all use neural models as part of the system, but their basic framework is still probabilistic, so they share the same problem of decision latency. In this paper, we focus on the \emph{pure} neural approach that uses neural models as the basic framework, so that short-time language information can be learned by frame-level discriminative training.
	
	%[1]  They investigated a rhythmic model, and?  CAn you add a second half to the sentence.  Did they conclude anything in particular?  It cuts off a little early.
	
	\subsection{Motivation of the paper}
	
	All the present neural LID methods are based on acoustic features, e.g., Mel filter banks (Fbanks) or Mel frequency cepstral coefficients (MFCCs), with phonetic information largely overlooked. This may have significantly hindered the performance of neural LID. Intuitively, it is a long-standing hypothesis that languages can be discriminated between by phonetic properties, either distributional or temporal; additionally, phonetic features represent information at a higher level than acoustic features, and so are more invariant with respect to noise and channels. Pragmatically, it has been demonstrated that phonetic information, either in the form of phone sequences, phone posteriors, or phonetic bottleneck features, can significantly improve LID accuracy in both the conventional PRLM approach~\cite{willmore2000comparing} and the more modern i-vector system~\cite{song2013vector,ferrer2016study,tian2016investigation}.  In this paper, we will investigate the utilization of phonetic information to improve neural LID. The basic concept is to use a phone-discriminative model to produce frame-level phonetic features, and then use these features to enhance RNN LID systems that were originally built with raw acoustic features.  The initial step is therefore feature combination, with the phonetic feature used as auxiliary information to assist acoustic RNN LID.  This is improved further, as additional research identified that a simpler model using only the phonetic feature as the RNN LID input provides even better performance. We call this RNN model based on phonetic features the \emph{phonetic temporal neural} LID approach, or PTN LID. As well as having a simplified model structure, the PTN offers deeper insight into the LID task by rediscovering the value of the phonetic temporal property in language discrimination. This property was historically widely and successfully applied in token-based approaches, e.g., PRLM~\cite{willmore2000comparing}, but has been largely overlooked due to the popularity of the i-vector approach.
	
	Table~\ref{tab:method2} summarizes different systems that use deep neural models in LID. The probabilistic approach uses DNNs as part of a probabilistic system, e.g., GMM or i-vector, while the neural approach uses various types of DNNs as the decision architecture. Both approaches may use either acoustic features or phonetic features. The proposed PTN approach is at the bottom-right of the table.

	\begin{table}[!t]
		\renewcommand{\arraystretch}{1}
		\caption{ LID methods with deep learning involvement.}
		\label{tab:method2}
		\centering
		\begin{tabular}{c|c|c}
			
			\hline
			& Probabilistic                                              & Neural  \\
			\hline
			Acoustic  & DNN i-vector~\cite{ferrer2016study,tian2016investigation}  & FFDNN~\cite{lopez2014automatic},RNN\cite{gonzalez2014automatic}    \\
			\hline
			Phonetic  & DNN feature i-vector~\cite{song2013vector}                 &  PTN (proposed) \\
			\hline
		\end{tabular}
	\end{table}

	\subsection{Paper organization}
	
	The remainder of the paper is organized as follows: the model structures of the PTN approach will be presented in Section~\ref{sec:arc}, which is followed by the implementation details in Section~\ref{sec:model}.  The experiments and results are reported in Section~\ref{sec:exp}, and some conclusions and future work will be presented in Section~\ref{sec:con}.

	\section{Phonetic neural modelling for LID}
	\label{sec:arc}
	
\begin{figure}[h]
		\centering
		\includegraphics[width=\linewidth]{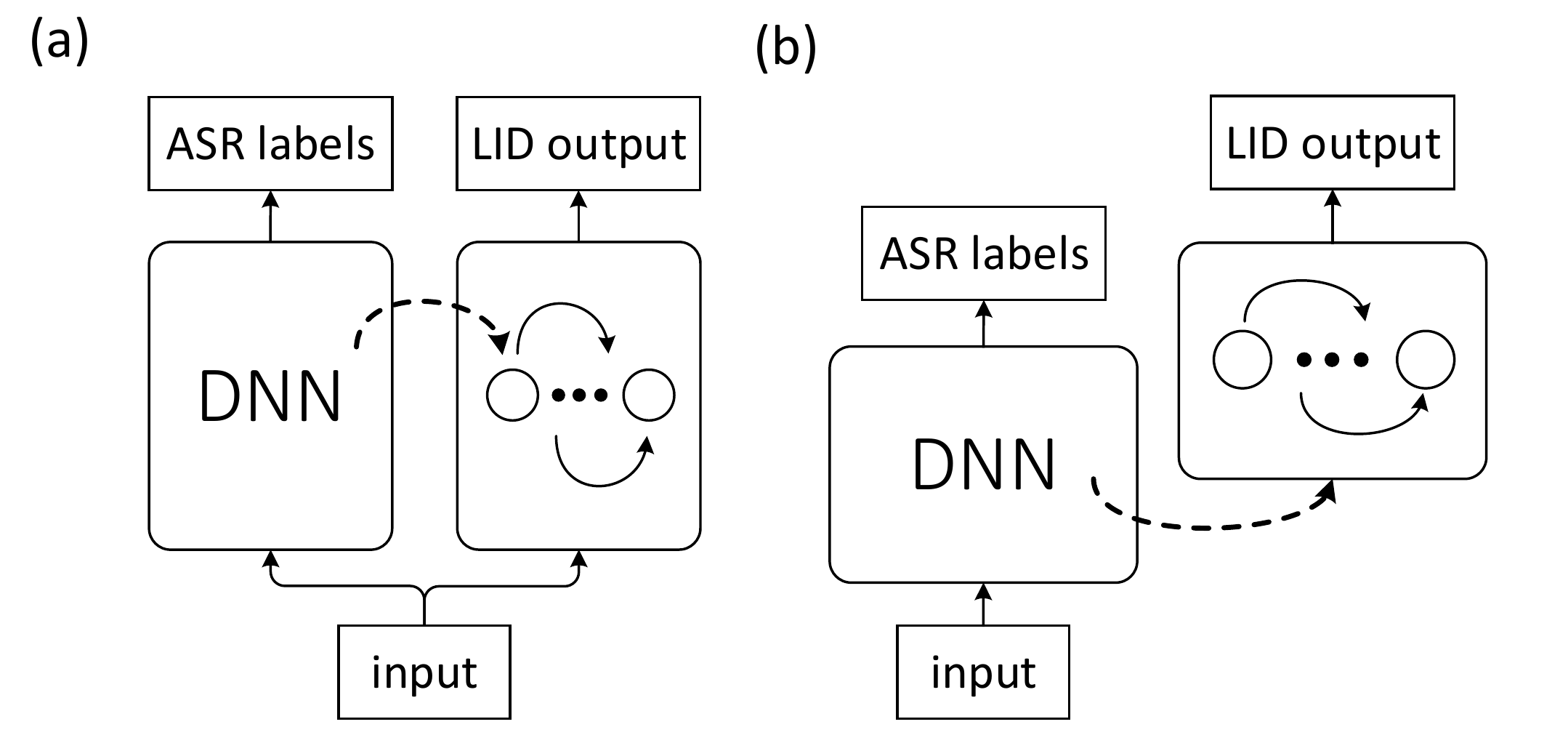}
		\caption{LID models employing phonetic information:
			(a) the phonetically aware model;
			(b) the PTN model.
			Both models consist of a phonetic DNN (left) to produce phonetic features and an LID RNN (right) to make LID decisions.
		}
		\label{fig:lid}
	\end{figure}

	%In this section, we present the models that employ phonetic information in the acoustic
	In this section, we present the models that employ phonetic information for RNN LID.  Although the phonetically aware approach treats phonetic information as auxiliary knowledge, the PTN approach uses phonetic information as the only input into the RNN LID system. Both are depicted in Fig.~\ref{fig:lid}.
	
	\subsection{Phonetically aware acoustic neural model}
	
	The instinctive idea for utilizing phonetic information in the RNN LID system is to treat it as auxiliary knowledge, which we call a phonetically aware approach.  Intuitively, this can be regarded as a knowledge-fusion method that uses both the phonetic and acoustic features to learn LID models. Fig.~\ref{fig:lid} (a) shows this model. A phonetic DNN model (this may be in any structure, such as FFDNN, RNN, TDNN) is used to produce frame-level phonetic features. These can be read from anywhere in the phonetic DNN, such as the output, or the last hidden layer, and then be propagated to the LID model, an LSTM-RNN in our study. This propagated phonetic information can be accepted by the LID model in different ways. For example, it can be part of the input, or as an additional term of the gate or non-linear activation functions.
	
	\subsection{Phonetic temporal neural model}
	
	The second model, which we call the PTN model, completely replaces the acoustic feature with the phonetic feature, and thus entirely relies on the properties of the phonetic representation. This learning is based on the RNN model, therefore the temporal patterns of the phonetic features can be learned.  This PTN system is shown in Fig.~\ref{fig:lid} (b). Although the PTN model is a special, `aggressive' case of the phonetically aware approach, the success of this model offers a deeper insight into the LID task as it rediscovers the importance of the temporal properties of phonetic representations.

	%\begin{figure}[h]
	%  \centering
	%  \includegraphics[width=0.8\linewidth]{fig/lid_2.pdf}
	%  \caption{The PTN LID architectures that consists a  phonetic DNN to produce phonetic features and
	%  an LID RNN model.}
	%  \label{fig:ptnm}
	%\end{figure}

	\subsection{Understanding the PTN approach}
	
	The rationality of the PTN approach can be understood from two perspectives: the phonetic perspective, which relates to what information is important, and the transfer learning perspective, which relates to how this information is learned.
	
	{\bf Phonetic perspective:} The PTN approach adopts the long-standing hypothesis (as used by the PRLM model) that languages should be discriminated by phonetic rather than spectral properties.  However this has been largely overlooked since the success of the i-vector approach, which achieved good performance using only raw acoustic features.  However, Song et al.~\cite{song2013vector} recently rediscovered the value of phonetic features in the i-vector model. The PTN approach proposed here follows the same idea and rediscovers the value of phonetic features in the neural model. We argue that this value is more important for the neural model than for the probabilistic model (e.g., i-vector), as its decision is based on only a small number of frames, and thus requires that the feature involves more language-related information and less noise and uncertainties. The i-vector model, in contrast, can utilize more speech signals, hence can discover language-related information from the distributional patterns even with raw acoustic features.
	
	Both the PTN approach and the historical token-based approach share the same idea of utilizing phonetic information and modelling the temporal patterns, but they are fundamentally different. Firstly, the phonetic information in the PTN approach is frame-level, while in conventional token-based methods this information is unit-level. Therefore, the PTN approach can represent phonetic properties at a higher temporal resolution. Secondly, conventional token-based methods represent phonetic information as sequences derived from phone recognition, while the PTN approach represents phonetic information as a feature vector that involves information contributed by all phones, and thus more detailed phonetic information is represented. Finally, the back-end model of the conventional token-based approach is an n-gram LM based on discrete tokens and trained with the maximum likelihood (ML) criterion, while the back-end model of the PTN approach is an RNN, which functions similarly to an RNN LM, but is based on continuous phonetic features, and trained with a task-oriented criterion that discriminates the target languages.
	
	%Note that the RNN-based LM was recently studied by Salamea et al.~\cite{salamea2016use}, though the input
	%is still phonetic units.

	{\bf Transfer learning perspective:} The second perspective to understand the PTN approach is from the transfer learning perspective~\cite{wang2015transfer}.  It is well known that DNNs perform very well at learning task-oriented features from raw data. This is the hypothesis behind conventional acoustic RNN LID methods: if the neural model is successfully trained, it can learn any useful information from the raw acoustic features layer by layer, including the phonetic information. It therefore initially seems unnecessary to design our PTN phonetic feature learning and modelling architecture.  However, we argue that using the language labels alone to learn LID-related information from raw acoustic features is highly ineffective, because these labels are too coarse to provide sufficient supervision.  With the PTN model, feature extraction is trained on speech data labelled with phones or words which are highly informative and fine-grained (compared to language labels), leading to a strong DNN model for phonetic feature extraction. Importantly, phone discrimination and language identification are naturally correlated (from our phonetic perspective), which means that the phonetic features learned with the strong phone/word supervision involves rich information suitable for LID. This is an example of transfer learning, where a related task (i.e., phone discrimination) is used to learn features for another task (LID).
	
	The PTN approach also involves another two transfer learning schemes: cross language and cross condition (databases).  This means that the phonetic DNN can be learned with any speech data in any language.  This property was identified in token-based LID~\cite{zissman1996comparison}, however it is more important for the phonetic neural models, as training the phonetic DNN requires a large amount of speech data which is often not available for the target languages and the operating conditions under test. Moreover, it is also possible to train the phonetic DNN with multilingual, multi-conditional data~\cite{huang2013cross}, resulting in robust and reliable phonetic feature extraction.
	
	In summary, the PTN approach utilizes a detailed phonetic representation (DNN phonetic feature), and a powerful temporal model (LSTM-RNN) to capture the phonetic temporal properties of a language with a high temporal resolution.  It also utilizes three types of transfer learning to ensure that the phonetic feature is representative and robust. Our PTN approach is therefore very powerful and flexible, and reconfirms the belief of many LID researchers that phonetic temporal information is highly valuable in language discrimination, not only for humans but also for machines.
	
	% \cite{lopez2014automatic} Automatic language identification using deep neural networks
	% \cite{gonzalez2014automatic} Automatic language identification using long short-term memory recurrent neural networks
	% \cite{ferrer2016study} Study of senone-based deep neural network approaches for spoken language recognition
	
	% \cite{song2013vector} using BN to replace MFCC or SDC to construct i-vector model
	% \cite{jin2016lid} {LID}-senone extraction via deep neural networks for end-to-end language identification}
	
	% \cite{lozano2015end} An end-to-end approach to language identification in short utterances using convolutional neural networks
	% \cite{kotov2016language} Language Identification Using Time Delay Neural Network D-Vector on Short Utterances
	% \cite{garcia2016stacked} Stacked Long-Term TDNN for Spoken Language Recognition
	
	% \cite{zazo2016language}Language identification in short utterances using long short-term memory ({LSTM}) recurrent neural networks
	% \cite{gelly2016divide} A Divide-and-Conquer Approach for Language Identification Based on Recurrent Neural Networks
	% \cite{salamea2016use} On the use of phone-gram units in recurrent neural networks for language identification, using RNNLM to score phonetic units
	% \cite{tian2016investigation} Investigation of Senone-based Long-Short Term Memory RNNs for Spoken Language Recognition, this
	%is i-vector based stuff
	
	\section{Model structure}
	\label{sec:model}
	
	This section presents the details of the phonetic neural LID models, including both the phonetically aware model and the PTN model. The phonetic DNN can be implemented in various DNN structures, and here we choose the TDNN~\cite{waibel1989phoneme} which can learn long-term phonetic patterns
	and performed well in our experiments.
	%Both models can learn long-term phonetic patterns and have shown great advantages for speech recognition~\cite{waibel1989phoneme,graves2013speech}.
	
	For the LID neural model, we choose the LSTM-RNN. One reason for this choice is that LSTM-RNN has been demonstrated to perform well in both the pure neural LID approach~\cite{gonzalez2014automatic} and the neural-probabilistic hybrid LID approach~\cite{tian2016investigation}.  Another reason is that the RNN model can learn the temporal properties of speech signals, which is in accordance with our motivation to model the phonetic dynamics, as in the conventional PRLM approach~\cite{matejka2006brno}. We first describe the LSTM-RNN structure used for LID, and then present the model structures of the phonetically aware acoustic RNN model and PTN model.
	
	\subsection{LSTM-RNN LID}
	
\begin{figure}[h]
		\centering
		\includegraphics[width=\linewidth]{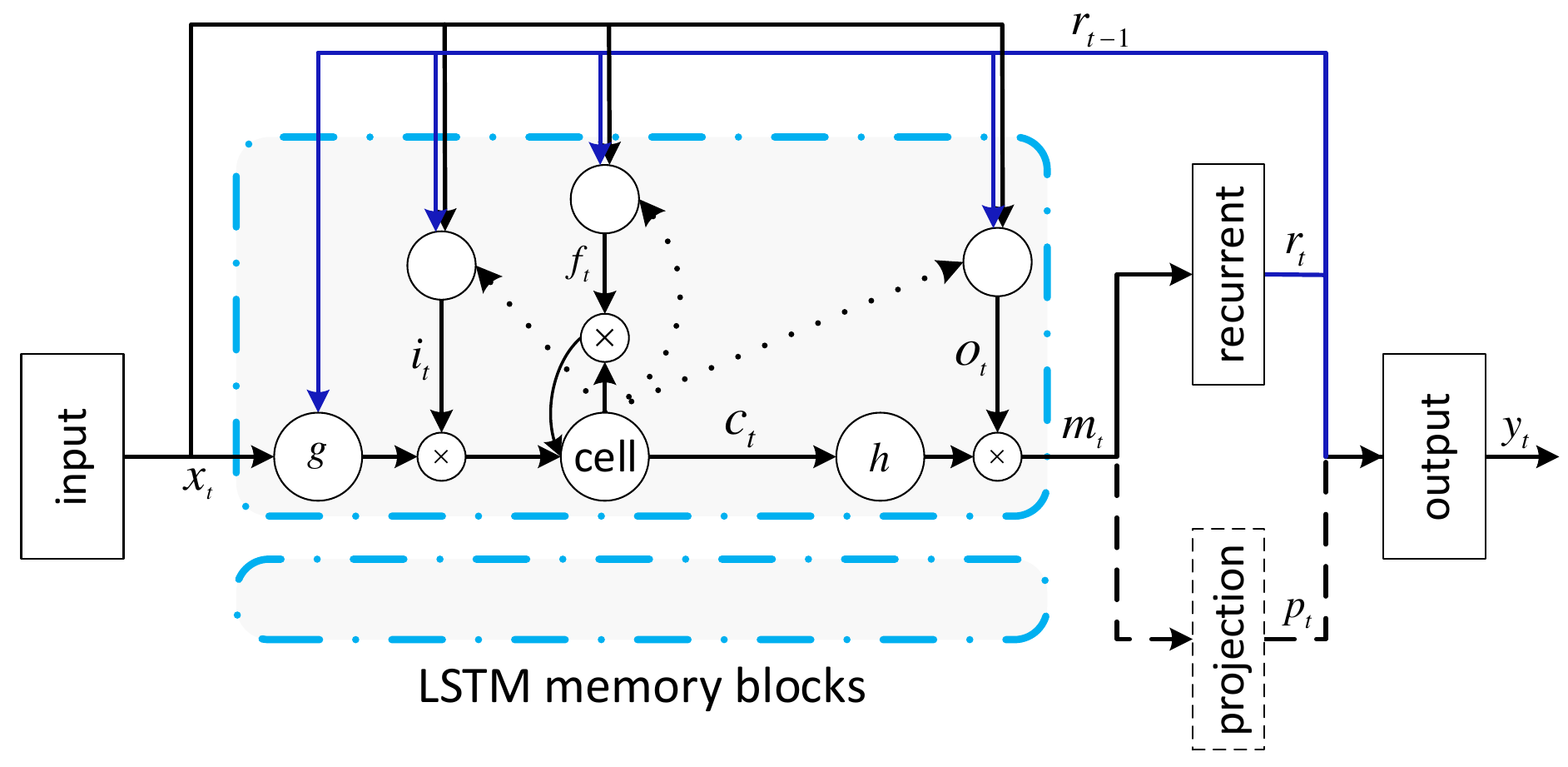}
		\caption{The LSTM model for the study. The picture is reproduced from~\cite{sak2014long}.}
		\label{fig:lstm}
	\end{figure}

	The LSTM-RNN model used in this study is a one-layer RNN model, where the hidden units are LSTM. The structure proposed by Sak et al.~\cite{sak2014long} is used, as shown in Fig.~\ref{fig:lstm}.

	The associated computation is given as follows:
	
	\begin{eqnarray}
	i_t &=& \sigma(W_{ix}x_{t} + W_{ir}r_{t-1} + W_{ic}c_{t-1} + b_i) \nonumber\\
	f_t &=& \sigma(W_{fx}x_{t} + W_{fr}r_{t-1} + W_{fc}c_{t-1} + b_f) \nonumber\\
	c_t &=& f_t \odot c_{t-1} + i_t \odot g(W_{cx}x_t + W_{cr}r_{t-1} + b_c) \nonumber\\
	o_t &=& \sigma(W_{ox}x_t + W_{or}r_{t-1} + W_{oc}c_t + b_o) \nonumber\\
	m_t &=& o_t \odot h(c_t) \nonumber\\
	r_t &=& W_{rm} m_t \nonumber\\
	p_t &=& W_{pm} m_t \nonumber\\
	y_t &=& W_{yr}r_t + W_{yp}p_t + b_y \nonumber
	\end{eqnarray}
	
	\noindent In the above equations, the $W$ terms denote weight matrices, and those associated with the cells were constrained to be diagonal in our implementation.  The $b$ terms denote bias vectors. $x_t$ and $y_t$ are the input and output symbols respectively; $i_t$, $f_t$, $o_t$ represent the input, forget and output gates, respectively; $c_t$ is the cell and $m_t$ is the cell output. $r_t$ and $p_t$ are two output components derived from $m_t$, where $r_t$ is recurrent and fed to the next time step, while $p_t$ is not recurrent and contributes to the present output only. $\sigma(\cdot)$ is the logistic sigmoid function, and $g(\cdot)$ and $h(\cdot)$ are non-linear activation functions, chosen to be hyperbolic. $\odot$ denotes element-wise multiplication.
	
	In this study, the LSTM layer consists of $1,024$ cells, and the dimensionality of both the recurrent and non-recurrent projections is set to $256$. The natural stochastic gradient descent (NSGD) algorithm~\cite{povey2014parallel} was employed to train the model. During the training and decoding, the cells were reset for each $20$ frames to ensure only short-time patterns are learned.
	
	\subsection{Phonetically aware neural LID}
	
	In the phonetically aware model, the phonetic feature is read from the phonetic DNN and is propagated to the LID RNN as additional information to assist the acoustic neural LID.  The phonetic feature can be read either from the output (phone posterior) or the last hidden layer (logits), and can be propagated to different components of the RNN LID model, e.g., the input/forget/output gates and/or
	the non-linear activation functions.
	
\begin{figure}[h]
		\centering
		\includegraphics[width=0.9\linewidth]{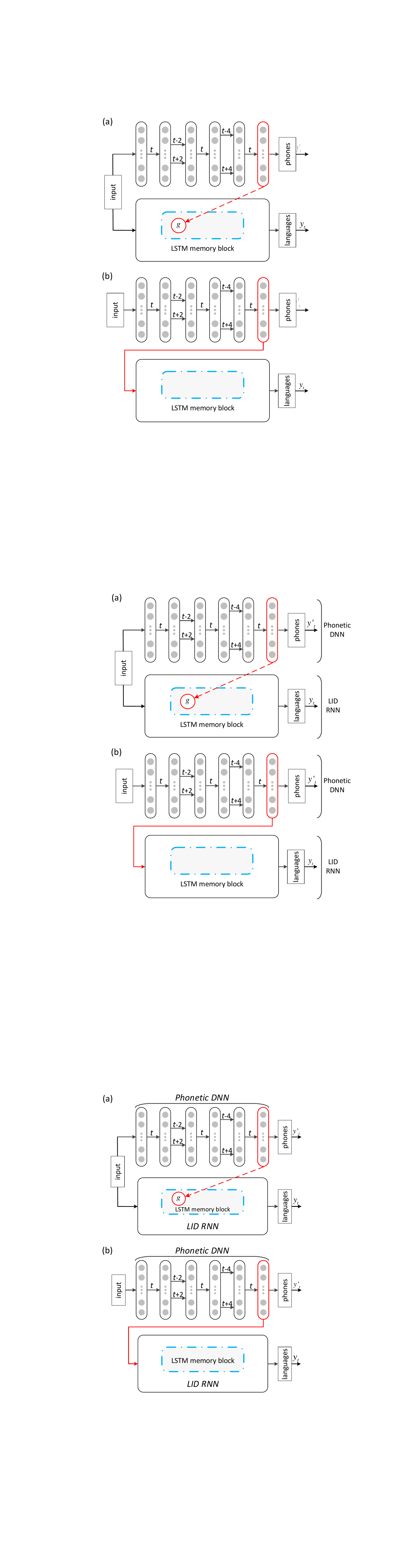}
		\caption{The phonetically aware RNN LID system (top) and the PTN LID system (bottom).
			The phonetic feature is read from the last hidden layer of the phonetic DNN which is a TDNN.
			The phonetic feature is then propagated to the $g$ function for the phonetically aware RNN LID system,
			and is the only input for the PTN LID system.}
		\label{fig:lid-eg}
	\end{figure}

	Fig.~\ref{fig:lid-eg} (a) illustrates a simple configuration, where the phonetic DNN is a TDNN model, and the feature is read from the
	last hidden layer. The phonetic feature is propagated to the non-linear function $g(\cdot)$.  With this configuration, calculation of the LID RNN is similar, except that the cell value should be updated as follows:
	
	\[
	c_t = f_t \odot c_{t-1} + i_t \odot g(W_{cx}x_t + W_{cr}r_{t-1} + \underline{W'_{c\phi}\phi_{t}} + b_c)
	\]
	
	\noindent where $\phi_t$ is the phonetic feature obtained from the phonetic DNN.
	
	\subsection{Phonetic temporal neural (PTN) LID}
	
	The phonetically aware acoustic RNN model is an acoustic-based approach, with the phonetic feature used as auxiliary information. In contrast, the PTN approach assumes that the phonetic temporal properties cover most of the information for language discrimination, so the acoustic feature is not important any more. Therefore, it removes all acoustic features and uses the phonetic features as the only input of the LID RNN, as shown in Fig.~\ref{fig:lid-eg} (b).
	
	%\begin{figure}[h]
	%  \centering
	%  \includegraphics[width=0.95\linewidth]{fig/lid-2.pdf}
	%  \caption{The PTN LID system.[change the picture, by replacing the phonetic DNN as TDNN]}
	%  \label{fig:ptnm}
	%\end{figure}
	
	It is interesting to compare the PTN approach with other LID approaches.  Firstly, it can be regarded as a new version of the conventional PRLM approach, particularly the recent PRLM implementation using RNN as the LM~\cite{salamea2016use}.  The major difference is that the PTN approach uses frame-level phonetic features while the PRLM approach uses token-level phonetic sequences; in addition, the phonetic information in the PTN approach is much richer than for PRLM, as it is represented as a continuous phonetic vector rather than discrete phonetic symbols.
	
	The PTN approach is also correlated to the neural-probabilistic hybrid approach, where the phonetic DNN is used to produce phonetic features, from which the GMM or i-vector model is constructed. The PTN approach uses the same phonetic features, but employs an RNN model to describe the dynamic property of the feature, instead of modelling the distributional property using GMM or i-vector models. As will be discussed in the next section, temporal modelling is very important for phonetic neural models.
	
	Finally, compared to the conventional acoustic RNN LID model, the PTN model uses phonetic features rather than acoustic features.
	Since the phonetic features can be learned with a very large speech database, they are much more robust against noise and uncertainties (e.g., speaker traits and channel distortions) than the raw acoustic features.  This suggests that the PTN approach is more robust against noise than the conventional acoustic RNN approach.

	\section{Experiments}
	\label{sec:exp}
	
	\subsection{Databases and configurations}
	
	The experiments were conducted on two databases: the Babel database and the AP16-OLR database.  The Babel database was collected as part of the IARPA (Intelligence Advanced Research Projects Activity) Babel program, which aimed to develop speech technologies for low-resource languages.  The sampling rate is $8$ kHz and the sample size is $16$ bits.  In this paper, we chose speech data from seven languages in the Babel database: Assamese, Bengali, Cantonese, Georgian, Pashto Tagalog and Turkish.  For each language, an official training and development dataset were provided.  The training datasets contain both conversational and scripted speech, and the development datasets only contain conversational speech.  We used the entire training set of each language for model training, but randomly selected $2,000$ utterances from the development set of each language to perform testing.
	
	The training data sets from the seven languages are as follows:
	Assamese  $75$ hours\footnote{Language collection release IARPA-babel102b-v0.5a.},
	Bengali $87$ hours\footnote{Language collection release IARPA-babel103b-v0.4b.},
	Cantonese $175$ hours\footnote{Language collection release IARPA-babel101b-v0.4c.},
	Georgian $64$ hours\footnote{Language collection release IARPA-babel404b-v1.0a.},
	Pashto $111$ hours\footnote{Language collection release IARPA-babel104b-v0.4bY.},
	Tagalog $116$ hours\footnote{Language collection release IARPA-babel106-v0.2g.}  and
	Turkish $107$ hours\footnote{Language collection release IARPA-babel105b-v0.5.}.
	The average duration of the test utterances is $4.15$ seconds, ranging from $0.19$ seconds to $30.85$
	seconds.
	
	The AP16-OL7 database was originally created by Speechocean Inc., targeted towards various speech processing tasks (mainly speech recognition),
	and was used as the official data for the AP16-OLR LID challenge\footnote{http://cslt.riit.tsinghua.edu.cn/mediawiki/index.php/OLR\_Challenge\_2016}.
	The database contains seven datasets, each in a particular language. These are: Mandarin, Cantonese, Indonesian, Japanese, Russian, Korean and Vietnamese. The data volume for each language is approximately $10$ hours of speech signals recorded by $24$ speakers ($12$ males and $12$ females), with each speaker recording approximately $300$ utterances in reading style by mobile phones, with a sampling rate of 16kHz and a sample size of 16 bits. Each dataset was split into a training set consisting of $18$ speakers, and a test set consisting of $6$ speakers. For Mandarin, Cantonese, Vietnamese and Indonesian, the recording was conducted in a quiet environment. For Russian, Korean and Japanese, there are $2$ recording conditions for each speaker, quiet and noisy.  The average duration (including silence) of all the $12,939$ test utterances of the seven languages is $4.74$ seconds, ranging from $1.08$ seconds to $18.06$ seconds.
	
	%{\color{red} Tabel here to describe the two databases.}
	
	The phonetic DNN is a TDNN structure, and the LID model is based on the LSTM-RNN. The raw feature used for those models consists of $23$-dimensional Fbanks, with a symmetric $2$-frame window for RNN and a symmetric $4$-frame window for TDNN
	to splice neighboring frames.  All the experiments were conducted with Kaldi~\cite{povey2011kaldi}. The default configurations of the Kaldi WSJ s5 nnet3 recipe were used to train the phonetic DNN and the LID RNN. We first report experiments based on the Babel database, and then experiments with the AP16-OLR database.
	
	\subsection{Babel: baseline of bilingual LID}
	\label{subsec:base_results}
	
	As the first step, we build three baseline LID systems, one based on the i-vector model, and the other two based on LSTM-RNN, using the speech data of two languages from Babel: Assamese and Georgian (AG).
	
	For the i-vector baseline, the UBM involves $2,048$ Gaussian components and the dimensionality of the i-vectors is $400$.  The static acoustic features consists of $12$-dimensional MFCCs and the log energy.  These static features are augmented by their first and second order derivatives, resulting in $39$-dimensional feature vectors.  In our experiment, we train an SVM for each language to determine the score of a test i-vector belonging to that language. The SVMs are trained on the i-vectors of all training segments, following the one-versus-rest strategy.
	
	The two RNN LID baselines are: a standard RNN LID system (AG-RNN-LID) that discriminates between the two languages in its output,
	and a multi-task system (AG-RNN-MLT) that was trained to discriminate between the two languages as well as the phones.  More precisely, the output units of the AG-RNN-MLT are separated into two groups: an LID group that involves two units corresponding to Assamese and Georgian respectively, and an ASR group that involves $3,349$ bilingual senones that are inherited from an HMM/GMM ASR system trained with the speech data of Assamese and Georgian, following the standard WSJ s5 HMM/GMM recipe of Kaldi.  The WSJ s5 nnet3 recipe of Kaldi is then used to train the AG-RNN-LID and AG-RNN-MLT systems.
	
	The LID task can be conducted by either AG-RNN-LID or AG-RNN-MLT (using the LID output group) at the frame-level (denoted as `Fr.'), using the frame-level language posteriors they produce. To evaluate the utterance-level (denoted as `Utt.') performance, the frame-level posteriors are averaged to form the utterance-level posterior, by which the language decision can be made.
	
	\begin{table}[thb!]
		\caption{\label{tab:base}{Results of the baseline LID systems for Babel AG.}}
		\centerline{
			\begin{tabular}{l|cc|cc}
				\hline
				& \multicolumn{2}{c|}{$C_{avg}$}  & \multicolumn{2}{c}{EER\%} \\
				\hline
				Model        &    Fr.       & Utt.    & Fr.       &  Utt. \\
				\hline
				i-vector     &    -         & 0.0893  & -         & 9.05  \\
				AG-RNN-LID   &    0.1545    & 0.0905  & 16.20     & 9.20  \\
				AG-RNN-MLT   &    0.0822    & 0.0399  & 8.68      & 4.10  \\
				%     AG-TD-MLT    & 0.1064       & 0.0355  & 11.17     & 3.75  \\
				\hline
			\end{tabular}
		}
	\end{table}
	
	The performance results with the three baseline systems, in terms of $C_{avg}$ and equal error rate (EER), are shown in Table~\ref{tab:base}.  The results indicate that both the LID RNN and the multi-task LID RNN are capable of language discrimination, and the multi-task RNN significantly outperforms both the LID RNN and the i-vector baseline.  This indicates that the phone information is very useful for neural LID, even if simply used as an auxiliary objective in the model training, hence supporting our transfer learning perspective, as described in Section~\ref{sec:arc}.
	
	The multi-task learning approach is an interesting way to involve phonetic information in LID.  However, it has the limitation of requiring the training data to be labelled in both languages and words/phones. This is very costly and not feasible in most scenarios.
	The phonetic neural models (the phonetically aware model and the PTN model) do not suffer from this problem.
	
	%[2]  I don't know what the one-verse-rest scheme is.  Is it common?  Do we need to add a sentence?
	%Answer: The spell is wrong. should be One-versus-rest

	\subsection{Babel: phonetically aware bilingual LID}
	\label{sec:phone-aware}

	The phonetically aware architecture uses phonetic features as auxiliary information to improve the RNN LID. We experimented with various architectures for the phonetic DNN, and found that the TDNN structure is a good choice. In this experiment, the TDNN structure is composed of $6$ time-delay layers, with each followed by a p-norm layer that reduces the dimensionality of the activation from $2,048$ to $256$, the same dimension as the recurrent layer of the LID LSTM-RNN. The activations of the last hidden layer in the TDNN are read out as the phonetic feature.
	
	Two TDNN models are trained. The AG-TDNN-MLT model is a multi-task model trained with the Assamese and Georgian data, and there are two groups of output targets, phone labels and language labels. The ASR performance (WER) of the AG-TDNN-MLT model is $66.4\%$ and $64.2\%$ for Assamese and Georgian respectively.  The SWB-TDNN-ASR model is an ASR model trained with the Switchboard database.  This database involves $317$ hours of telephone speech signals in English, recorded from $4,870$ speakers.  The ASR performance (WER) of SWB-TDNN-ASR is $20.8\%$ on the Eval2000 dataset.
	
	\begin{table}[thb!]
		\caption{\label{tab:phone-bi-lid}{Results of phonetically aware RNN LID for Babel AG.}}
		\centerline{
			\begin{tabular}{l|l|cc|cc}
				\hline
				&               & \multicolumn{2}{c|}{$C_{avg}$}  & \multicolumn{2}{c}{EER\%}\\
				\hline
				Phonetic DNN     & Info. Rec.    & Fr.         & Utt.      & Fr.      & Utt. \\
				\hline
				%      AG-RNN-MLT       & input gate    & 0.0762      & 0.0298    &  7.99    & 3.25    \\
				%      AG-RNN-MLT       & forget gate   & 0.0765      & 0.0280    &  8.01    & 3.20    \\
				%      AG-RNN-MLT       & output gate   & 0.0752      & 0.0286    &  7.91    & 3.15   \\
				%      AG-RNN-MLT       & $g$ function  & \textbf{0.0726}  & \textbf{0.0264}  & \textbf{7.57}  & \textbf{2.92}    \\
				%      \hline
				%      AG-RNN-ASR       & output gate   & 0.0801      & 0.0298    &  8.38    & 3.28    \\
				%      AG-RNN-ASR       & $g$ function  & 0.0795      & 0.0283    &  8.38    & 3.08    \\
				%      A-RNN-ASR        & output gate   & 0.0942      & 0.0375    &  9.81   & 4.10     \\
				%      A-RNN-ASR        & $g$ function  & 0.0900      & 0.0380    &  9.49   & 4.05     \\
				%      SWB-RNN-ASR      & output gate   & 0.0969      & 0.0438    &  10.07   & 4.6     \\
				%      SWB-RNN-ASR      & $g$ function  & 0.0970      & 0.0385    &  10.18   & 4.10    \\
				%      \hline
				%      AG-RNN-LID       & output gate   & 0.1310      & 0.0638    &  13.67   & 7.00   \\
				%      AG-RNN-LID       & $g$ function  & 0.1279      & 0.0640    &  13.45   & 6.95   \\
				%      \hline
				%      AG-TDNN-MLT      & -             & 0.1064      & 0.0355	 &  11.17   & 3.75     \\ (what is this???)
				AG-TDNN-MLT      & $g$ function  & \textbf{0.0435}  & \textbf{0.0214}  & \textbf{4.45}  & \textbf{2.40}     \\
				SWB-TDNN-ASR     & $g$ function  & 0.1029       & 0.0355     & 10.58    & 3.95     \\
				\hline
			\end{tabular}
		}
	\end{table}
	
	Another design decision that had to be made was to choose which component in the LID RNN will receive the phonetic information. After a series of preliminary experiments, it was found that the $g$ function is the best receiver. With this choice and the two TDNN phonetic DNNs, we therefore build the phonetically aware LID system. The results are shown in Table~\ref{tab:phone-bi-lid}. Several conclusions can be obtained from the results.
	
	\begin{itemize}
		\item The phonetically aware system significantly outperforms the baseline RNN LID system (second row of the results in Table~\ref{tab:base}).  This suggests that involving phonetic information with RNN LID has clear benefits.
		
		\item The phonetically aware system significantly outperforms the multi-task RNN LID (third row of the results in Table~\ref{tab:base}). Note that in the multi-task RNN LID, the phonetic knowledge is used as an auxiliary task to assist the LID RNN training and has
		shown great benefits. The advantages of the phonetically aware system demonstrated that using the phonetic knowledge to produce phonetic features seems to be a better method than using the knowledge to directly assist model training.
		
		\item The phonetic DNN trained with Assamese and Georgian data (AG-TDNN-MLT) shows better  performance than the one trained  with the Switchboard dataset (SWB-TDNN-ASR).  This is not surprising as Assamese and Georgian are the two languages chosen to discriminate between in the experiments presented in this section, so AG-TDNN-MLT is more consistent with this LID task. Nevertheless, it is still highly interesting to observe that clear benefits can be obtained by using phonetic features produced by SWB-TDNN-ASR, which is trained with a completely irrelevant dataset, in terms of both languages and environmental conditions. This confirmed our transfer learning perspective theory (as discussed previously), and demonstrated that phonetic features are largely portable and the phonetic DNN can be trained with any data in any languages. This observation is particularly interesting for LID tasks on low-resource languages, as the phonetic DNN can be trained with data from any rich-resource languages.
	\end{itemize}

	%[3]
	%[3]  I discussed this section via email.  However, I would also like to add that I don't think you've defined switchboard speech anywhere.

	\subsection{Babel: PTN for bilingual LID}
	
	In the above experiments, the phonetic feature is used as auxiliary information. Here, we evaluate the PTN architecture where the phonetic feature entirely replaces the acoustic features (Fbanks).  The experiment is conducted with two phonetic DNN models: AG-TDNN-MLT and SWB-TDNN-ASR.
	
	The results are presented in Table~\ref{tab:tdnn-phone-lid}.  We first observe that the PTN systems perform as well as the best phonetically aware system in Table~\ref{tab:phone-bi-lid}, and even better in terms of the utterance-level EER.  For better comparison, we also test the special case of the phonetically aware RNN LID (Ph. Aware), where both the phonetic and acoustic features are used as the LID RNN input (Ph+Fb).  This is the same as the PTN model, but involves additional acoustic features.  The results are shown in the second group of Table~\ref{tab:tdnn-phone-lid}. It can be seen that this feature combination does not provide any notable improvement to the results.  This means that the phonetic feature is sufficient to represent the distinctiveness of each language, in accordance with our argument that language characters are mostly phonetic.
	
	\begin{table}[thb!]
		\caption{\label{tab:tdnn-phone-lid}{Results of PTN LID and phonetically aware RNN LID with both phonetic and acoustic features for Babel AG.}}
		\resizebox{\columnwidth}{!}{
			\centerline{
				\begin{tabular}{l|l|c|cc|cc}
					\hline
					&               &                    &\multicolumn{2}{c|}{$C_{avg}$} & \multicolumn{2}{c}{EER\%} \\
					\hline
					System     & Phonetic DNN  & LID Feature        & Fr.          & Utt.       & Fr.      & Utt. \\
					\hline
					PTN        & AG-TDNN-MLT   & Phonetic           & 0.0436       & 0.0205     & 4.49     & 2.33     \\
					PTN        & SWB-TDNN-ASR  & Phonetic           & 0.1042       & 0.0355     & 10.77    & 3.88     \\
					\hline
					Ph. Aware  & AG-TDNN-MLT   & Ph+Fb              & 0.0440       & 0.0213     & 4.57     & 2.33        \\
					Ph. Aware  & SWB-TDNN-ASR  & Ph+Fb              & 0.1014       & 0.0348     & 10.53    & 4.00     \\
					\hline
				\end{tabular}
			}
		}
	\end{table}

	We also attempted to use the TDNN as the LID model (replacing the RNN) to learn static (rather than temporal) patterns of the phonetic features. We found that this model failed to converge.  The same phenomenon was also observed in the AP16-OLR experiment (which will be discussed later in the paper).  This is an important observation and it suggests that, with the phonetic feature, only the temporal properties are informative for language discrimination.
	
	\subsection{Babel: Phonetic knowledge or deep structure?}
	
	The good performance using only the phonetic features (i.e. the PTN approach) leads to the question of how this performance advantage in comparison to the RNN LID baseline is obtained.  This paper has discussed the phonetic and transfer learning perspectives, which jointly state that the main advantage of PTN is the phonetic knowledge learned through transfer learning. However, another possible reason is that the deeper architecture consisting of both the phonetic DNN and the LID RNN may help to learn more abstract features. If the latter reason is more important, than a similar deep structure with only the LID labels can work similarly well. To answer this question, we design the following three experiments to test the contributions to the results from phonetic information (transfer learning) and deep architecture (deep learning):
	
	\begin{itemize}
		
		\item TDNN-LSTM. The phonetic DNN, TDNN in the experiment, is initialized randomly and trained together with the LID RNN. This means
		that the TDNN is not trained with ASR labels, but as part of the LID neural model, and is trained end-to-end.
		
		\item Pre-trained TDNN-LSTM. The same as TDNN-LSTM, except that the TDNN is initialized by AG-TDNN-MLT.
		
		\item 3-layer LSTM-RNN. The 1-layer LSTM-RNN LID model may be not strong enough to learn useful information from acoustic features, hence leading to the suboptimal performance in Table~\ref{tab:base}. We experiment with a 3-layer LSTM-RNN LID system to test if a simple deeper network can obtain the same performance as with the phonetic feature.
		
	\end{itemize}
	
	The results of these three deep models are shown in Table~\ref{tab:acoustic-lid}. The TDNN-LSTM model completely fails.  Using the phonetic TDNN as the initialization helps the training, but the results are worse than directly using the phonetic model. This means that the phonetic feature is almost optimal, and does not require any further LID-oriented end-to-end training. Finally, involving more LSTM layers (3-layer LSTM-RNN) does improve the performance a little when compared to the one-layer LSTM baseline ($7.70$ vs $9.20$, ref. to Table~\ref{tab:base}).  %This is consistent with our observation in Table~\ref{tab:phone-bi-lid}, where using a LID-trained phonetic DNN (approximately equal to 2-layer LSTM layers but with the two layers trained layer-by-layer) improves results slightly.
	These results indicate that the improvement with the PTN architecture is mainly due to the phonetic information it has learned from the ASR-oriented training (sometimes by multi-task learning), rather than the deep network structure.  In other words, it is the transfer learning instead of deep learning that improves LID performance with the PTN architecture.

	\begin{table}[thb!]
		\caption{\label{tab:acoustic-lid}{Results of deeper LID models for Babel AG.}}
		\centerline{
			\begin{tabular}{l|cc|cc}
				\hline
				& \multicolumn{2}{c|}{$C_{avg}$}  & \multicolumn{2}{c}{EER\%}\\
				\hline
				LID Model              & Fr.          & Utt.       & Fr.      & Utt. \\
				\hline
				TDNN-LSTM              & 0.5000       & 0.5000     & 58.14    & 50.00     \\
				Pre-trained TDNN-LSTM   & 0.0500       & 0.0234     & 5.69     & 2.55     \\
				3-layer LSTM-RNN       & 0.1415       & 0.0750     & 14.82    & 7.70     \\
				\hline
			\end{tabular}
		}
	\end{table}

	\subsection{Babel: PTN on seven languages}
	
	We evaluate various LID models on the seven languages of the Babel database.  First, the i-vector and LSTM-RNN LID baselines are presented.  For the i-vector system, linear discriminative analysis (LDA) is employed to promote language-related information before training SVMs. The dimensionality of the LDA projection space is set to $6$.  For the phonetically aware RNN and the PTN systems, two phonetic DNNs are evaluated, AG-TDNN-MLT and SWB-TDNN-ASR. For the phonetically aware system, the $g$ function of the LSTM-RNN LID model is chosen as the receiver.  The results are shown in Table~\ref{tab:phone-babel}.  It can be seen that both the phonetically aware and the PTN systems outperform the i-vector baseline and the acoustic RNN LID baseline, and that the PTN system with the AG-TDNN-MLT phonetic DNN performs the best. The SWB-TDNN-ASR performs slightly worse than AG-TDNN-MLT, indicating that familiarity with the language and the environment is beneficial when discriminating between languages. However, phonetic DNNs trained with data in foreign languages and in mismatched environment conditions (e.g., SWB-TDNN-ASR) still work well.

	\begin{table*}[thb!]
		\caption{\label{tab:phone-babel}{Results of various LID systems on the $7$ languages in Babel.}}
		\centerline{
			\begin{tabular}{l|l|c|c|c|cc|cc}
				\hline
				&              &           &            &             & \multicolumn{2}{c|}{$C_{avg}$} & \multicolumn{2}{c}{EER\%}\\
				\hline
				System        & Phonetic DNN & LID model & Info. Rec. & LID Feature & Fr.  & Utt.    & Fr.    & Utt. \\
				\hline
				i-vector      & -            & i-vector  & -          & -        & -       & 0.1696  & -      & 16.52  \\
				Acoustic RNN  & -            & LSTM-RNN  & -          & Fbank    & 0.1987  & 0.1249  & 19.11  & 12.63  \\
				\hline
				Ph. Aware     & AG-TDNN-MLT  & LSTM-RNN  & $g$        & Fbank    & 0.1280  & 0.0620  & 12.27  & 6.66 \\
				Ph. Aware     & SWB-TDNN-ASR & LSTM-RNN  & $g$        & Fbank    & 0.1610  & 0.0786  & 15.49  & 8.24  \\
				\hline
				PTN           & AG-TDNN-MLT  & LSTM-RNN  & Input      & Phonetic & \textbf{0.1165}  & \textbf{0.0518}  & \textbf{11.12}   & \textbf{5.70} \\
				PTN           & SWB-TDNN-ASR & LSTM-RNN  & Input      & Phonetic & 0.1726  & 0.0823  & 16.65  & 8.56  \\
				\hline
			\end{tabular}
		}
	\end{table*}

	%[4]  Did you mean to say "Channel" here?
	%Changed a little
	
	\subsection{AP16-OLR: PTN on seven languages }
	
	In this section, we test the phonetic RNN LID approach on the AP16-OLR database. Compared to the Babel database, the speech signals in AP16-OLR are broadband (sampling rate of 16k Hz), and the acoustic environment is less noisy. Additionally, the speech data of each language is much more limited (10 hours per language), so we assume that training a phonetic DNN model is not feasible with the data of the target languages. We therefore utilize transfer learning, i.e., using phonetic DNNs trained on data in other languages.
	
	All the test conditions are the same as in the $7$ language Babel experiment. We trained two phonetic DNNs: one is a TDNN model of the same size as the AG-TDNN-ASR model in Section~\ref{sec:phone-aware}, but trained on the WSJ database, denoted by `WSJ-TDNN-ASR'.  The other is also a TDNN, but is taken from an industry project, trained on a speech database involving $10,000$ hours of Chinese speech signals with $40$ dimensional Fbanks. The network contains $7$ rectifier TDNN layers, each containing $1,200$ hidden units. This model is denoted by `CH-TDNN-ASR'.  The weight matrix of the last hidden layer in CH-TDNN-ASR is decomposed by SVD, where
	the low rank is set to $400$.  The $400$-dimensional activations are read from the low-rank layer and are used as the phonetic feature.
	
	The test results on the seven languages in the database are shown in Table~\ref{tab:phone-ap16}. It can be seen that the phonetic RNN LID models, either the phonetically aware RNN or the PTN approach, significantly outperform the acoustic RNN baseline system. The PTN system seems much more effective, which differs from the Babel database results. This may be attributed to the limited training data, so the simpler PTN architecture is preferred.  Comparing the WSJ-based phonetic DNN and the Chinese phonetic DNN, the Chinese model is better. This may be attributed to several reasons: (1) the Chinese database contains a larger volume of training data; (2) Chinese is one of the seven languages in AP16-OLR; (3) Chinese is more similar to the remaining $6$ target languages in comparison to English, as most of the languages in AP16-OLR are oriental languages.

	\begin{table*}[thb!]
		\caption{\label{tab:phone-ap16}{Results of various LID systems on the $7$ languages in AP16-OLR.}}
		\centerline{
			\begin{tabular}{l|l|c|c|c|cc|cc}
				\hline
				&     &            &&& \multicolumn{2}{c|}{$C_{avg}$} & \multicolumn{2}{c}{EER\%}\\
				\hline
				System       & Phonetic DNN  & LID Model & Info. Rec. & LID Feature            & Fr. & Utt. & Fr. & Utt. \\
				\hline
				i-vector     & -             & i-vector  & -      &           & -       & \textbf{0.0383}  & -  & \textbf{3.49}  \\
				Acoustic RNN & -             & LSTM-RNN  & -      & Fbank     & 0.2467  & 0.1983  & 25.70 & 20.33  \\
				\hline
				Ph. Aware    & WSJ-TDNN-ASR  & LSTM-RNN  & $g$    & Fbank     & 0.1857  & 0.1183  & 18.99 & 11.96   \\
				Ph. Aware    & CH-TDNN-ASR   & LSTM-RNN  & $g$    & Fbank     & 0.1527  & 0.1109  & 15.77 & 11.91  \\
				\hline
				Ph. Aware    & WSJ-TDNN-ASR  & LSTM-RNN  & Input  & Ph+Fb     & 0.1816  & 0.1107  & 18.47 & 11.33    \\
				Ph. Aware    & CH-TDNN-ASR   & LSTM-RNN  & Input  & Ph+Fb     & 0.1153  & 0.0591  & 11.93 & 6.95   \\
				\hline
				PTN          & WSJ-TDNN-ASR  & LSTM-RNN  & Input  & Phonetic  & 0.1683  & 0.0697  & 17.08 & 7.57   \\
				PTN          & CH-TDNN-ASR   & LSTM-RNN  & Input  & Phonetic  & \textbf{0.1126}  & \textbf{0.0524}  & \textbf{11.74} & \textbf{6.34}  \\
				\hline
			\end{tabular}
		}
	\end{table*}
	
	Another observation is that the i-vector system outperforms the phonetic RNN systems in the AP16-OLR experiment, which is inconsistent with the observations in the Babel experiment, where both the phonetic systems, significantly outperform the i-vector system.  This discrepancy can be attributed to the different data profiles of the two databases, with two possible key factors: (1) the utterances of AP16-OLR are longer than Babel, making the i-vector system more effective; (2) the speech signals of AP16-OLR are cleaner than those of Babel. The RNN system is more robust against noise, and this advantage is less prominent with clean data. We will examine the two conjectures in the following experiments.
	
	\subsection{AP16-OLR: utterance duration effect}
	
	To show the relative advantage of the RNN and the i-vector systems on utterances of different length, we select the utterances of at least $5$ seconds from the AP16-OLR test set, and create $10$ test sets by dividing them into small utterances of different durations, from $0.5$ seconds to $5$ seconds, in steps of $0.5$ seconds.  Each group contains $5,907$ utterances, and each utterance in a group is a
	random segment excerpted from the original utterance.
	
\begin{figure}[h]
		\centering
		\includegraphics[width=0.95\linewidth]{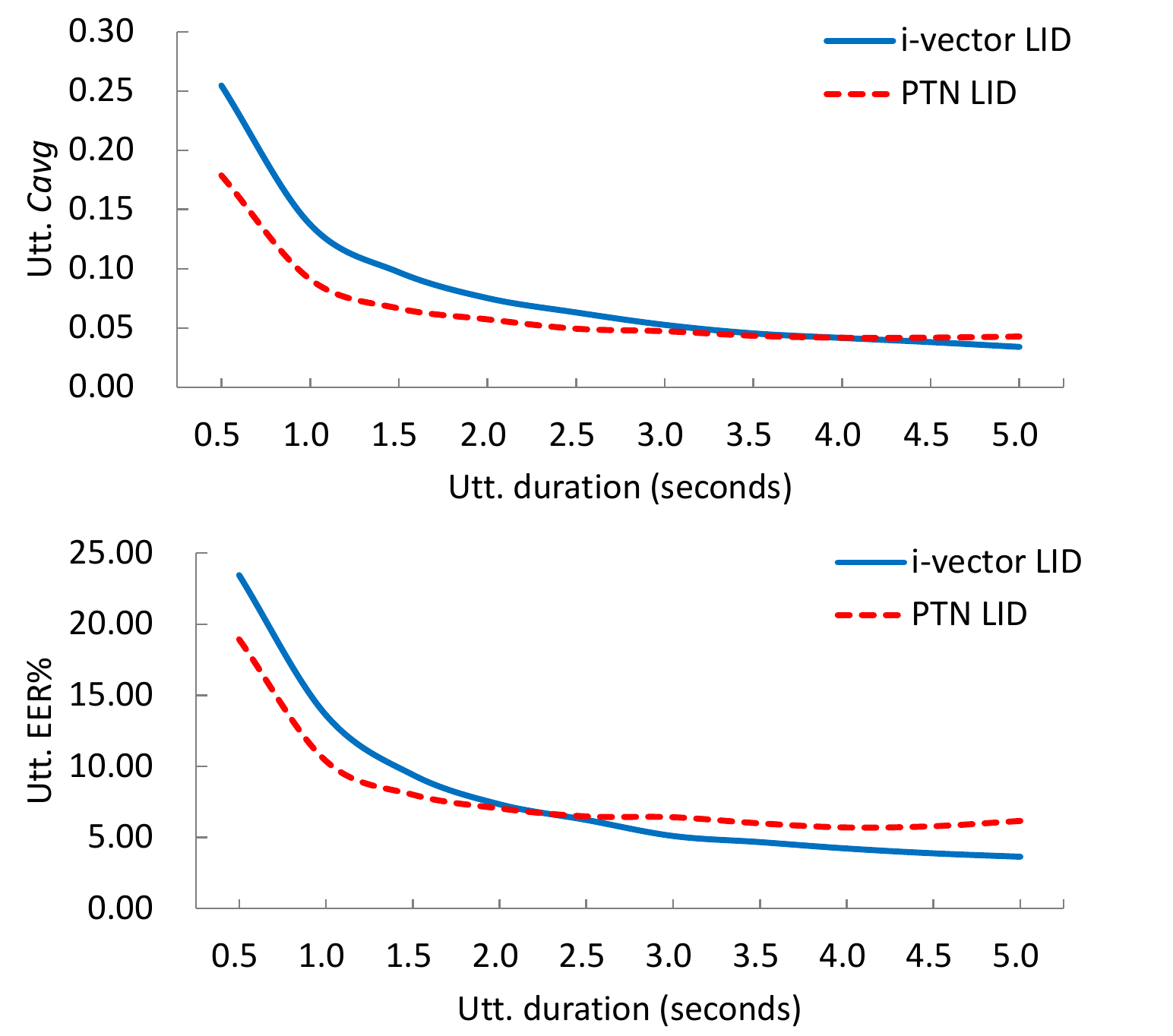}
		\caption{Comparison of the effect of utterance duration on i-vector LID and PTN LID in terms of utterance $C_{avg}$ (above) and EER (bottom).}
		\label{fig:duration}
	\end{figure}

	The performance of the i-vector and PTN systems on the $10$ test sets are shown in Fig.~\ref{fig:duration}, in terms of $C_{avg}$ and EER respectively. It is clear that the PTN system is more effective on short utterances, and if the utterance duration is more than $3$ seconds, the i-vector system is the best performer, especially in terms of EER.
	
\begin{figure}[h]
		\centering
		\includegraphics[width=\linewidth]{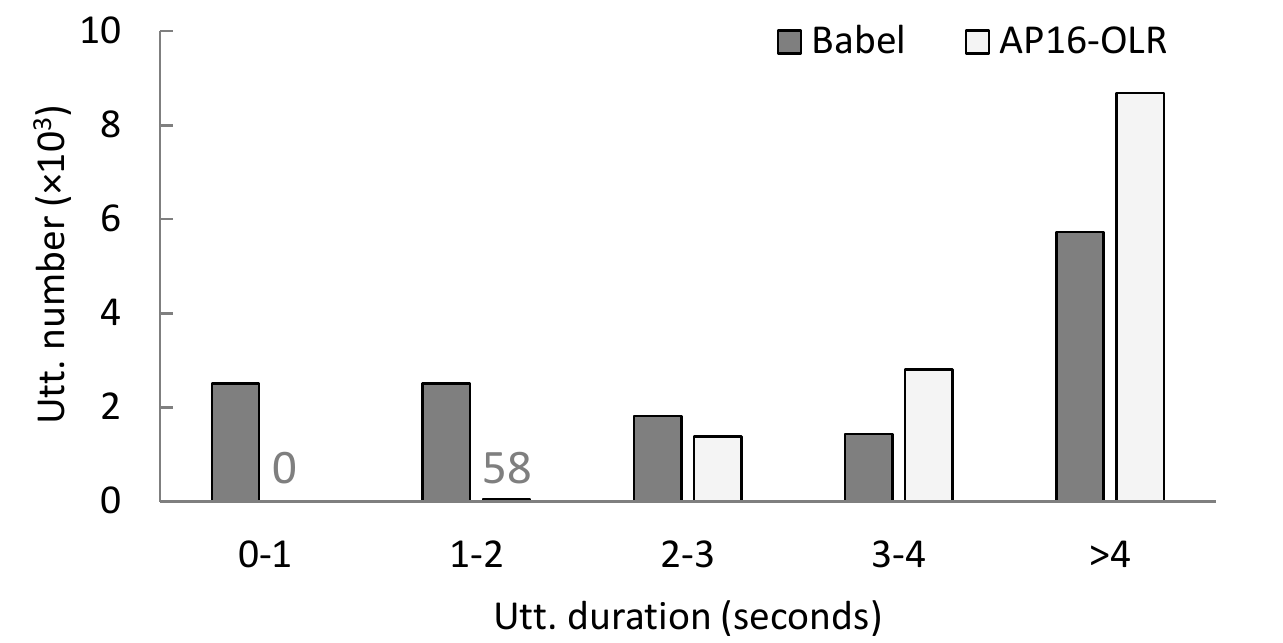}
		\caption{Duration distribution of the test utterances of the Babel database and the AP16-OLR database.}
		\label{fig:dur-dist}
	\end{figure}

	The duration distribution of the test utterances of the Babel database and the AP16-OLR database are shown in Fig.~\ref{fig:dur-dist}. It is clear that the test utterances are generally longer in AP16-OLR than in Babel. This explains why the relative performance of the i-vector system and the RNN system is inconsistent between the two databases.
	
	\subsection{AP16-OLR: noise robustness}
	
\begin{figure}[!h]
		\centering
		\includegraphics[width=\linewidth]{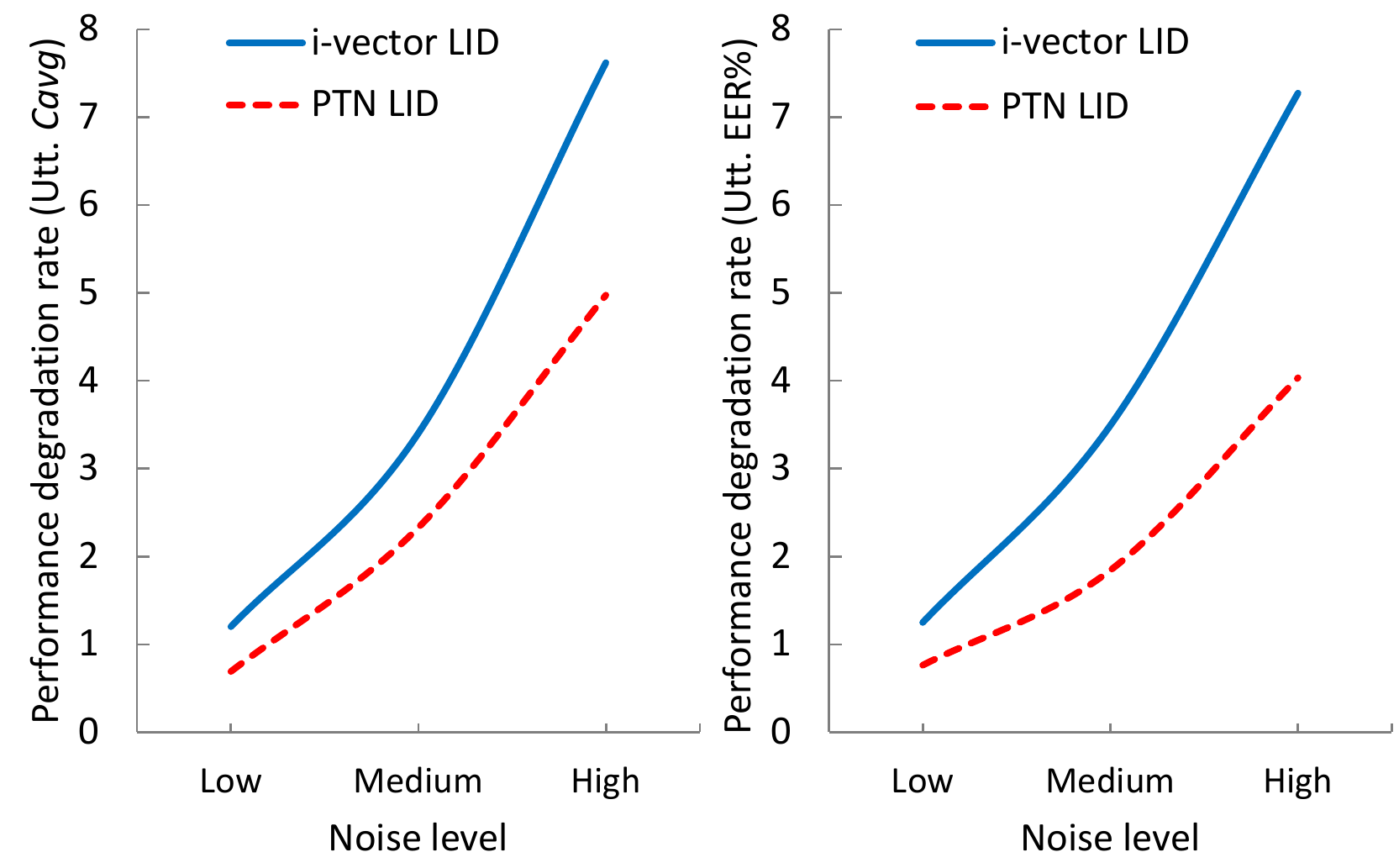}
		\caption{Performance degradation rate of i-vector LID and PTN LID with noise in terms of utterance $C_{avg}$ (left) and EER (right).}
		\label{fig:noise}
	\end{figure}

	Finally, we test the hypothesis that the RNN system is more robust against noise.  Firstly white noise is added to the AP16-OLR test set at different SNR levels, and the noise-augmented data are tested on two systems: the i-vector baseline and the best performing PTN system from Table~\ref{tab:phone-ap16}, i.e. with CH-TDNN-ASR as the phonetic DNN.  The results of these two systems with different
	levels of white noise are shown in Table~\ref{tab:lid-white}.  It can be seen that the PTN system is more noise-robust: with more noise corruption, the gap between the i-vector system and the PTN system becomes less significant, and the PTN system is better than the i-vector system in terms of $C_{avg}$ when the noise level is high (SNR=$10$).  This can be observed more clearly in Fig.~\ref{fig:noise}, where the performance degradation rates compared to the noise-free condition are shown. The figure shows that when the noise increases, the performance degradation with the PTN system is less significant compared to the degradation with the i-vector system.  As the Babel speech data is much more noisy than the AP16-OLR speech, this noise robustness with the PTN approach partly explains why the relative performance is inconsistent between the two databases.

	\begin{table}[thb!]
		\caption{\label{tab:lid-white}{Results of i-vector LID and PTN LID with different levels of noise.}}
		\centerline{
			\begin{tabular}{c|cc|cc|cc|cc}
				\hline
				& \multicolumn{4}{c|}{i-vector LID}   & \multicolumn{4}{c}{PTN LID}    \\
				\hline
				& \multicolumn{2}{c|}{$C_{avg}$}  & \multicolumn{2}{c|}{EER\%}  & \multicolumn{2}{c|}{$C_{avg}$}  & \multicolumn{2}{c}{EER\%} \\
				\hline
				SNR (dB)  &  Fr.     & Utt.    & Fr.   & Utt.     &  Fr.     & Utt.    & Fr.   & Utt. \\
				\hline
				30        &  -       & 0.0843  & -      & 7.84    &  0.1762  & 0.0886  & 18.32 & 11.16  \\
				20        &  -       & 0.1686  & -      & 15.67   &  0.2435  & 0.1744  & 24.68 & 18.01  \\
				10        &  -       & 0.3300  & -      & 28.87   &  0.3489  & 0.3129  & 34.41 & 31.90 \\
				\hline
			\end{tabular}
		}
	\end{table}

	%\subsection{Discussion}
	%More languages.
	%More complex models.

	\section{Conclusions}
	\label{sec:con}
	
	This paper proposed a phonetic temporal neural (PTN) approach for language identification.  In this approach, acoustic features are substituted for phonetic features to build an RNN LID model. Our experiments conducted on the Babel and  AP16-OLR databases demonstrated that the PTN approach can provide dramatic performance improvement over the baseline RNN LID system, with even better results than a phonetically aware approach that treats the phonetic feature as additional auxiliary information.  This demonstrated that phonetic temporal information is much more informative
	than raw acoustic information for discriminating between languages. This was a long-standing belief of LID researchers in the PRLM era, but has been doubted since the increased popularity and utilization of the i-vector approach in recent years.  Future work will improve the performance of the neural LID approach on long sentences, by enabling the LSTM-RNN to learn long-time patterns, e.g., by multi-scale RNNs~\cite{chung2016hierarchical}.

	%\section{Acknowledgements}
	%\newpage
	\bibliographystyle{IEEEtran}
	\bibliography{deeplid}

\begin{thebibliography}{10}
\providecommand{\url}[1]{#1}
\csname url@samestyle\endcsname
\providecommand{\newblock}{\relax}
\providecommand{\bibinfo}[2]{#2}
\providecommand{\BIBentrySTDinterwordspacing}{\spaceskip=0pt\relax}
\providecommand{\BIBentryALTinterwordstretchfactor}{4}
\providecommand{\BIBentryALTinterwordspacing}{\spaceskip=\fontdimen2\font plus
\BIBentryALTinterwordstretchfactor\fontdimen3\font minus
  \fontdimen4\font\relax}
\providecommand{\BIBforeignlanguage}[2]{{%
\expandafter\ifx\csname l@#1\endcsname\relax
\typeout{** WARNING: IEEEtran.bst: No hyphenation pattern has been}%
\typeout{** loaded for the language `#1'. Using the pattern for}%
\typeout{** the default language instead.}%
\else
\language=\csname l@#1\endcsname
\fi
#2}}
\providecommand{\BIBdecl}{\relax}
\BIBdecl

\bibitem{fromkin2010introduction}
V.~Fromkin, R.~Rodman, and N.~Hyams, \emph{An introduction to language}.\hskip
  1em plus 0.5em minus 0.4em\relax Cengage Learning, 2010.

\bibitem{comrie2009world}
B.~Comrie, \emph{The world's major languages}.\hskip 1em plus 0.5em minus
  0.4em\relax Routledge, 2009.

\bibitem{crystal2cambridge}
D.~Crystal, \emph{The Cambridge encyclopedia of language}.\hskip 1em plus 0.5em
  minus 0.4em\relax Cambridge University Press, 2010.

\bibitem{haper07}
M.~P. Harper and M.~Maxwell, ``Spoken language characterization,'' in
  \emph{Springer Handbook of Speech Processing}.\hskip 1em plus 0.5em minus
  0.4em\relax Springer, 2008, pp. 797--810.

\bibitem{muthusamy1994perceptual}
Y.~K. Muthusamy, N.~Jain, and R.~A. Cole, ``Perceptual benchmarks for automatic
  language identification,'' in \emph{Proceedings of IEEE International
  Conference on Acoustics, Speech and Signal Processing (ICASSP)},
  vol.~1.\hskip 1em plus 0.5em minus 0.4em\relax IEEE, 1994, pp. 333--336.

\bibitem{mori1999human}
K.~Mori, N.~Toba, T.~Harada, T.~Arai, M.~Komatsu, M.~Aoyagi, and Y.~Murahara,
  ``Human language identification with reduced spectral information.'' in
  \emph{Proceedings of European Conference On Speech, Communication and
  Technology (EUROSPEECH)}, 1999, pp. 391--394.

\bibitem{navratil2001spoken}
J.~Navratil, ``Spoken language recognition-a step toward multilinguality in
  speech processing,'' \emph{IEEE Transactions on Speech and Audio Processing},
  vol.~9, no.~6, pp. 678--685, 2001.

\bibitem{cimarusti1982development}
D.~Cimarusti and R.~Ives, ``Development of an automatic identification system
  of spoken languages: Phase i,'' in \emph{Proceedings of IEEE International
  Conference on Acoustics, Speech and Signal Processing (ICASSP)},
  vol.~7.\hskip 1em plus 0.5em minus 0.4em\relax IEEE, 1982, pp. 1661--1663.

\bibitem{foil1986language}
J.~Foil, ``Language identification using noisy speech,'' in \emph{Proceedings
  of IEEE International Conference on Acoustics, Speech and Signal Processing
  (ICASSP)}, vol.~11.\hskip 1em plus 0.5em minus 0.4em\relax IEEE, 1986, pp.
  861--864.

\bibitem{torres2002approaches}
P.~A. Torres-Carrasquillo, E.~Singer, M.~A. Kohler, R.~J. Greene, D.~A.
  Reynolds, and J.~R. Deller~Jr, ``Approaches to language identification using
  gaussian mixture models and shifted delta cepstral features.'' in
  \emph{Proceedings of the Annual Conference of International Speech
  Communication Association (INTERSPEECH)}, 2002, pp. 89--92.

\bibitem{zissman1993automatic}
M.~A. Zissman, ``Automatic language identification using gaussian mixture and
  hidden markov models,'' in \emph{Proceedings of IEEE International Conference
  on Acoustics, Speech and Signal Processing (ICASSP)}, vol.~2.\hskip 1em plus
  0.5em minus 0.4em\relax IEEE, 1993, pp. 399--402.

\bibitem{willmore2000comparing}
J.~Willmore, R.~Price, and W.~Roberts, ``Comparing gaussian mixture and neural
  network modelling approaches to automatic language identification of
  speech,'' in \emph{Australasian International Conference on Speech Science
  and Technology (SST)}, 2000, pp. 74--77.

\bibitem{wong2004automatic}
K.~Wong and M.-h. Siu, ``Automatic language identification using discrete
  hidden markov model.'' in \emph{Proceedings of the Annual Conference of
  International Speech Communication Association (INTERSPEECH)}, 2004, pp.
  1633--1636.

\bibitem{nakagawa1992speaker}
S.~Nakagawa, Y.~Ueda, and T.~Seino, ``Speaker-independent, text-independent
  language identification by hmm.'' in \emph{International Conference on Spoken
  Language Processing (ICSLP)}, vol.~92, 1992, pp. 1011--1014.

\bibitem{kwasny1992identifying}
S.~C. Kwasny, B.~L. Kalman, W.~Wu, and A.~M. Engebretson, ``Identifying
  language from speech: An example of high-level, statistically-based feature
  extraction,'' in \emph{Proceedings of the Annual Conference of the Cognitive
  Science Society (CogSci)}, 1992, pp. 909--914.

\bibitem{muthusamy1993segmental}
Y.~K. Muthusamy, ``A segmental approach to automatic language identification,''
  Ph.D. dissertation, Jawaharlal Nehru Technological University, 1993.

\bibitem{campbell2004language}
W.~M. Campbell, E.~Singer, P.~A. Torres-Carrasquillo, and D.~A. Reynolds,
  ``Language recognition with support vector machines,'' in \emph{Proceedings
  of Odyssey}, 2004, pp. 41--44.

\bibitem{Najim2011lang}
N.~Dehak, A.-C. Pedro, D.~Reynolds, and R.~Dehak, ``Language recognition via
  i-vectors and dimensionality reduction,'' in \emph{Proceedings of the Annual
  Conference of International Speech Communication Association (INTERSPEECH)},
  2011, pp. 857--860.

\bibitem{martinez2011language}
D.~Mart{\i}nez, O.~Plchot, L.~Burget, O.~Glembek, and P.~Matejka, ``Language
  recognition in ivectors space,'' in \emph{Proceedings of the Annual
  Conference of International Speech Communication Association (INTERSPEECH)},
  2011, pp. 861--864.

\bibitem{zissman1996comparison}
M.~A. Zissman \emph{et~al.}, ``Comparison of four approaches to automatic
  language identification of telephone speech,'' \emph{IEEE Transactions on
  speech and audio processing}, vol.~4, no.~1, pp. 31--44, 1996.

\bibitem{matejka2006brno}
P.~Matejka, L.~Burget, P.~Schwarz, and J.~Cernocky, ``Brno university of
  technology system for nist 2005 language recognition evaluation,'' in
  \emph{IEEE Odyssey Speaker and Language Recognition Workshop}.\hskip 1em plus
  0.5em minus 0.4em\relax IEEE, 2006, pp. 1--7.

\bibitem{hazen1997segment}
T.~J. Hazen and V.~W. Zue, ``Segment-based automatic language identification,''
  \emph{The Journal of the Acoustical Society of America}, vol. 101, no.~4, pp.
  2323--2331, 1997.

\bibitem{zhu2005different}
D.~Zhu, M.~Adda-Decker, and F.~Antoine, ``Different size multilingual phone
  inventories and context-dependent acoustic models for language
  identification.'' in \emph{Proceedings of the Annual Conference of
  International Speech Communication Association (INTERSPEECH)}, 2005, pp.
  2833--2836.

\bibitem{schultz1996lvcsr}
T.~Schultz, I.~Rogina, and A.~Waibel, ``Lvcsr-based language identification,''
  in \emph{Proceedings of IEEE International Conference on Acoustics, Speech
  and Signal Processing (ICASSP)}, vol.~2.\hskip 1em plus 0.5em minus
  0.4em\relax IEEE, 1996, pp. 781--784.

\bibitem{hieronymus1997robust}
J.~L. Hieronymus and S.~Kadambe, ``Robust spoken language identification using
  large vocabulary speech recognition,'' in \emph{Proceedings of IEEE
  International Conference on Acoustics, Speech and Signal Processing
  (ICASSP)}, vol.~2.\hskip 1em plus 0.5em minus 0.4em\relax IEEE, 1997, pp.
  1111--1114.

\bibitem{rouas2003modeling}
J.-L. Rouas, J.~Farinas, F.~Pellegrino, and R.~Andr{\'e}-Obrecht, ``Modeling
  prosody for language identification on read and spontaneous speech,'' in
  \emph{Proceedings of IEEE International Conference on Acoustics, Speech and
  Signal Processing (ICASSP)}, vol.~6.\hskip 1em plus 0.5em minus 0.4em\relax
  IEEE, 2003, pp. 40--43.

\bibitem{lopez2014automatic}
I.~Lopez-Moreno, J.~Gonzalez-Dominguez, O.~Plchot, D.~Martinez,
  J.~Gonzalez-Rodriguez, and P.~Moreno, ``Automatic language identification
  using deep neural networks,'' in \emph{Proceedings of IEEE International
  Conference on Acoustics, Speech and Signal Processing (ICASSP)}.\hskip 1em
  plus 0.5em minus 0.4em\relax IEEE, 2014, pp. 5337--5341.

\bibitem{lozano2015end}
A.~Lozano-Diez, R.~Zazo~Candil, J.~Gonz{\'a}lez~Dom{\'\i}nguez, D.~T. Toledano,
  and J.~Gonzalez-Rodriguez, ``An end-to-end approach to language
  identification in short utterances using convolutional neural networks,'' in
  \emph{Proceedings of the Annual Conference of International Speech
  Communication Association (INTERSPEECH)}, 2015, pp. 403--407.

\bibitem{jin2016lid}
M.~Jin, Y.~Song, I.~Mcloughlin, L.-R. Dai, and Z.-F. Ye, ``{LID}-senone
  extraction via deep neural networks for end-to-end language identification,''
  in \emph{Proceedings of Odyssey}, 2016, pp. 210--216.

\bibitem{kotov2016language}
M.~Kotov and M.~Nastasenko, ``Language identification using time delay neural
  network d-vector on short utterances,'' in \emph{International Conference on
  Speech and Computer}, vol. 9811.\hskip 1em plus 0.5em minus 0.4em\relax
  Springer, 2016, pp. 443--449.

\bibitem{garcia2016stacked}
D.~Garcia-Romero and A.~McCree, ``Stacked long-term tdnn for spoken language
  recognition,'' in \emph{Proceedings of the Annual Conference of International
  Speech Communication Association (INTERSPEECH)}, 2016, pp. 3226--3230.

\bibitem{gonzalez2014automatic}
J.~Gonzalez-Dominguez, I.~Lopez-Moreno, H.~Sak, J.~Gonzalez-Rodriguez, and
  P.~J. Moreno, ``Automatic language identification using long short-term
  memory recurrent neural networks.'' in \emph{Proceedings of the Annual
  Conference of International Speech Communication Association (INTERSPEECH)},
  2014, pp. 2155--2159.

\bibitem{gelly2016divide}
G.~Gelly, J.-L. Gauvain, V.~Le, and A.~Messaoudi, ``A divide-and-conquer
  approach for language identification based on recurrent neural networks,'' in
  \emph{Proceedings of the Annual Conference of International Speech
  Communication Association (INTERSPEECH)}, 2016, pp. 3231--3235.

\bibitem{zazo2016language}
R.~Zazo, A.~Lozano-Diez, J.~Gonzalez-Dominguez, D.~T.~Toledano, and
  J.~Gonzalez-Rodriguez, ``Language identification in short utterances using
  long short-term memory (lstm) recurrent neural networks,'' \emph{PLOS ONE},
  vol.~11, pp. 1--17, 2016.

\bibitem{song2013vector}
Y.~Song, B.~Jiang, Y.~Bao, S.~Wei, and L.-R. Dai, ``I-vector representation
  based on bottleneck features for language identification,'' \emph{Electronics
  Letters}, vol.~49, no.~24, pp. 1569--1570, 2013.

\bibitem{ferrer2016study}
L.~Ferrer, Y.~Lei, M.~McLaren, and N.~Scheffer, ``Study of senone-based deep
  neural network approaches for spoken language recognition,'' \emph{IEEE/ACM
  Transactions on Audio, Speech and Language Processing}, vol.~24, no.~1, pp.
  105--116, 2016.

\bibitem{tian2016investigation}
Y.~Tian, L.~He, Y.~Liu, and J.~Liu, ``Investigation of senone-based long-short
  term memory rnns for spoken language recognition,'' in \emph{Proceedings of
  Odyssey}, 2016, pp. 89--93.

\bibitem{wang2015transfer}
D.~Wang and T.~F. Zheng, ``Transfer learning for speech and language
  processing,'' in \emph{Proceedings of Asia-Pacific Signal and Information
  Processing Association Annual Summit and Conference (APSIPA)}.\hskip 1em plus
  0.5em minus 0.4em\relax IEEE, 2015, pp. 1225--1237.

\bibitem{huang2013cross}
J.-T. Huang, J.~Li, D.~Yu, L.~Deng, and Y.~Gong, ``Cross-language knowledge
  transfer using multilingual deep neural network with shared hidden layers,''
  in \emph{Proceedings of IEEE International Conference on Acoustics, Speech
  and Signal Processing (ICASSP)}.\hskip 1em plus 0.5em minus 0.4em\relax IEEE,
  2013, pp. 7304--7308.

\bibitem{waibel1989phoneme}
A.~Waibel, T.~Hanazawa, G.~Hinton, K.~Shikano, and K.~J. Lang, ``Phoneme
  recognition using time-delay neural networks,'' \emph{IEEE Transactions on
  Acoustics, Speech, and Signal Processing}, vol.~37, no.~3, pp. 328--339,
  1989.

\bibitem{sak2014long}
H.~Sak, A.~W. Senior, and F.~Beaufays, ``Long short-term memory recurrent
  neural network architectures for large scale acoustic modeling,'' in
  \emph{Proceedings of the Annual Conference of International Speech
  Communication Association (INTERSPEECH)}, 2014, pp. 338--342.

\bibitem{povey2014parallel}
D.~Povey, X.~Zhang, and S.~Khudanpur, ``Parallel training of deep neural
  networks with natural gradient and parameter averaging,'' \emph{arXiv
  preprint arXiv:1410.7455}, 2014.

\bibitem{salamea2016use}
C.~Salamea, L.~F. D'Haro, R.~de~C{\'o}rdoba, and R.~San-Segundo, ``On the use
  of phone-gram units in recurrent neural networks for language
  identification,'' in \emph{Proceedings of Odyssey}, 2016, pp. 117--123.

\bibitem{povey2011kaldi}
D.~Povey, A.~Ghoshal, G.~Boulianne, L.~Burget, O.~Glembek, N.~Goel,
  M.~Hannemann, P.~Motlicek, Y.~Qian, and P.~Schwarz, ``The kaldi speech
  recognition toolkit,'' in \emph{Proceedings of IEEE 2011 workshop on
  Automatic Speech Recognition and Understanding}, no. EPFL-CONF-192584.\hskip
  1em plus 0.5em minus 0.4em\relax IEEE Signal Processing Society, 2011.

\bibitem{chung2016hierarchical}
J.~Chung, S.~Ahn, and Y.~Bengio, ``Hierarchical multiscale recurrent neural
  networks,'' \emph{arXiv preprint arXiv:1609.01704}, 2016.

\end{thebibliography}
	
\end{document}